\documentclass{article}
\usepackage{spconf,amsmath,cite}
\usepackage{graphicx}
\usepackage{color}

\usepackage{multirow}

\usepackage{amsfonts,amsmath,amssymb,amsthm, amsopn}

\newcommand{\norm}[1]{\left\lVert#1\right\rVert} 
\newcommand{\vassilis}[1]{{\color{red}#1}}

\usepackage{setspace}

\DeclareMathOperator{\simty}{sim}

\usepackage{enumitem}
\setlist[itemize]{itemsep=0pt, topsep=5pt}
\setlist[enumerate]{itemsep=0pt, topsep=0pt}
\setlist[enumerate]{noitemsep, topsep=0pt}

\title{Song recommendation with Non-Negative Matrix factorization and graph total variation}

\name{Kirell Benzi, Vassilis Kalofolias, Xavier Bresson and Pierre Vandergheynst}
\address{Signal Processing Laboratory 2 (LTS2), Swiss Federal Institute of Technology (EPFL)}

\begin{document}

\maketitle

\begin{abstract}
This work formulates a novel song recommender system\footnote{The code is available at: \href{https://github.com/kikohs/recog}{https://github.com/kikohs/recog}}as a matrix completion problem that benefits from collaborative filtering through Non-negative Matrix Factorization (NMF) and content-based filtering via total variation (TV) on graphs. The graphs encode both playlist proximity information and song similarity, using a rich combination of audio, meta-data and social features. As we demonstrate, our hybrid recommendation system is very versatile and incorporates several well-known methods while outperforming them.
Particularly, we show on real-world data that our model overcomes w.r.t. two evaluation metrics the recommendation of models solely based on low-rank information, graph-based information or a combination of both.

\end{abstract}


\vspace{-6pt}
\section{Introduction}\label{sec:introduction}
\vspace{-6pt}
Recommending movies on Netflix, friends on Facebook, or jobs on LinkedIn are tasks gaining an increasing interest over the last years. Low-rank matrix factorization techniques \cite{koren2009matrix} where amongst the winners of the famous Netflix prize, involving explicit user ratings as input. Similar techniques were soon used in order to solve implicit feedback problems, where item preferences were implied for example by the actions of a user \cite{Hu08collaborativefiltering, pro:RendleFreudenthalerGantnerSchmidt09Playlist}.
Specifically regarding songs and playlists recommendation, various techniques have been proposed, ranging from pure content-based methods \cite{van2013deep} to hybrid models \cite{shao2009music}. A comprehensive review of related algorithms can be found in \cite{celma2010, Bonnin:2014ej}.
Recently, graph regularization was proposed in order to enhance the quality of matrix completion problems \cite{ma2011recommender, art:KalofoliasBressonBronsteinVandergheynst14MCGraphs, cai2011graph}. 

\noindent
The contributions of this paper are as follows:
\begin{itemize}
\item A mathematically sound hybrid system that benefits from collaborative and content-based filtering.
\item The introduction of a new graph regularization term (TV) \cite{art:RudinOsherFatemi92ROF} in the context of recommendation that outperforms the widely used Tikhonov regularization. \cite{cai2011graph, art:KalofoliasBressonBronsteinVandergheynst14MCGraphs},
\item A well-defined iterative optimization scheme based on proximal splitting methods \cite{art:CombettesPesquet11ProxReview}.
\end{itemize}
Numerical experiments demonstrate the performance of our proposed recommender system.

\begin{figure}[t]
\centering
\includegraphics[height=5cm]{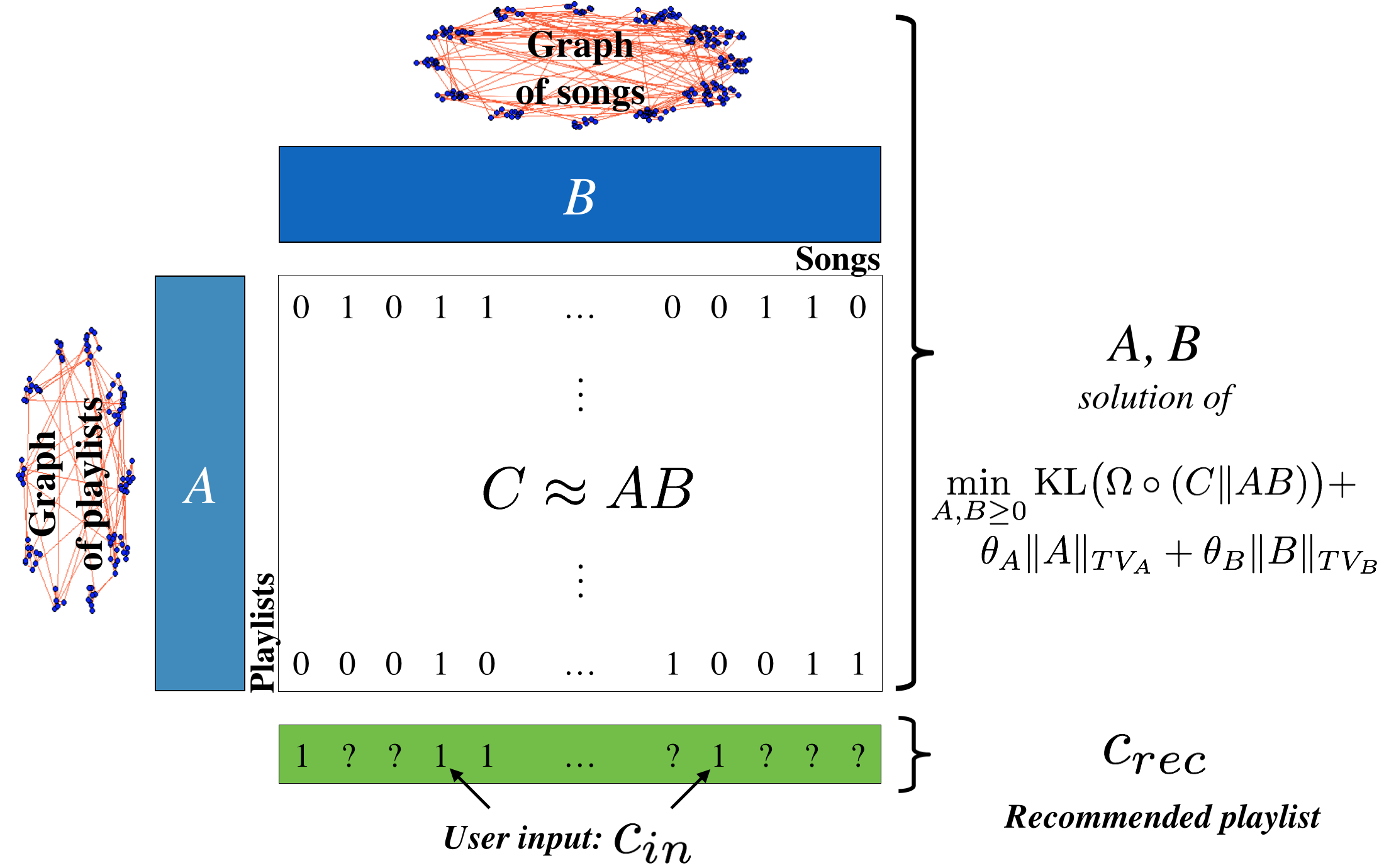}  
\caption{The architecture of our playlist recommender system.}
\label{fig_recom}
\end{figure}

\vspace{-6pt}
\section{Our Recommendation Algorithm}\label{sec:models}
\vspace{-6pt}

Suppose we are given $n$ playlists, each containing some of $m$ songs. We define matrix $C\in\{0,1\}^{n \times m}$ as in \cite{pro:HaririMobasherBurke12Playlist,pro:RendleFreudenthalerGantnerSchmidt09Playlist}, that has a value $C_{ij} = 1$ if playlist $i$ contains song $j$, $0$ otherwise. We also define a weight mask $\Omega \in\{\varepsilon, 1\}^{n \times m}$ that has a ''confidence'' value $\Omega_{ij}=1$ one if the entry $C_{ij}$ is $1$, and a small value $\varepsilon$, otherwise (we use $\varepsilon = 0.1$). This follows the example of implicit feedback problems \cite{Hu08collaborativefiltering}, since a zero in matrix $C$ does not mean that the corresponding song is irrelevant to the playlist, but that it is less probably relevant. 

The goal of the training step is to find an approximate low-rank representation $AB \approx C$, where $A\in\mathbb{R}^{n \times r}_+$, $B\in\mathbb{R}^{r \times m}_+$ non-negative and with small $r$. This problem is known as Non-Negative Matrix Factorization (NMF) and has drawn a lot of attention after the seminal work \cite{lee1999learning}. 
The advantage of NMF over other factorization techniques is that the approximation is only based on adding factors, a property explained as \textit{learning the parts of objects} \cite{lee1999learning}, in this case the playlists. NMF comes to the cost of being NP-hard \cite{vavasis2009complexity}, so sophisticated regularization is important for finding a good local minimum. In our problem we use outside information given by the songs and playlists graphs to give structure to the factors $A$ and $B$. Our model is formulated as
\begin{eqnarray}
\label{our_model}
\min_{A,B\geq 0}\textrm{KL}  \big( \Omega \circ ( C\|AB) \big) + \theta_A \|A\|_{TV_A} + \theta_B \|B\|_{TV_B}, \!\!\!
\end{eqnarray}
where $\circ$ is the pointwise multiplication operator and $\theta_A, \theta_B\in\mathbb{R}_+$. We use a weighted Kullback-Leibler (KL) divergence as a distance measure between $C$ and $AB$, that has been shown to be more accurate than the Frobenius norm for various NMF settings \cite{art:YanezBach14NMF}. 
The second term is the TV of the rows of $A$ on the playlists graph, so penalizing it promotes piecewise constant signals \cite{art:RudinOsherFatemi92ROF}. Similarly with the third term for columns of $B$.
Eventually, the proposed model leverages the works of \cite{art:KalofoliasBressonBronsteinVandergheynst14MCGraphs,art:YanezBach14NMF}, and extends them to graphs using the TV semi-norm. 

\noindent\\[-7pt]\textbf{Graph Regularization with Total Variation.}
\vassilis{}
In our NMF-based recommender, each playlist $i$ is represented in a low-dimensional space by a row $A_i$ of the matrix $A$. In order to learn better low-rank representations $A_i$ of the playlists, we also impose the pairwise similarities of the playlists $w_{ii'}^A$ on their corresponding low-rank representations. We can see this from the definition of the TV regularization term, $\|A\|_{TV_A} = \frac{1}{2}\sum_i\sum_{i'\sim i}w_{ii'}^A\|A_i-A_{i'}\|_1$. Hence, when two playlists $i,i'$ are similar then they are also well-connected on the graph and the weight of the edge connecting these two playlists $w_{ii'}^A$ is large (here $w_{ii'}^A\approx 1$). Moreover, any large distance between the corresponding low-dimensional representation vectors $(A_i,A_{i'})$ is penalized, forcing $(A_i,A_{i'})$ to stay close in the low-dimensional space. In a similar way, each song $j$ is represented in a low-dimensional space by a column $B_j$ of the matrix $B$. If two songs $(j,j')$ are close ($w_{jj'}^B\approx 1$), so will be $(B_j,B_{j'})$ with the graph regularization $\|B\|_{TV_B}$. 

A similar idea has been used in \cite{cai2011graph} by incorporating the graph information through Tikhonov regularization, i.e.  with the Dirichlet energy term $\frac{1}{2}\sum_i\sum_{i'\sim i}w_{ii'}^A\|A_i-A_{i'}\|_2^2$. However, the latter promotes smooth changes between the columns of $A$, while the graph TV term penalization promotes piecewise constant signals with potentially sharp transitions between columns $A_i$ and $A_{i'}$. This is advantageous in applications where well separated classes are sought, for example in clustering \cite{art:BressonLaurentUminskyVonBrecht13MTV}, or in our recommendation system where similar playlists might belong to different categories.

As we demonstrate in Sec.~\ref{sec:results}, the use of the graphs of songs and playlists improve significantly the recommendations, while the results are better when the more forgiving TV term is used instead of Tikhonov regularization.

\noindent\\[-7pt]
\textbf{Primal-dual optimization.} Optimization problem \eqref{our_model} is globally non-convex, but separately convex w.r.t. $A$ and $B$. A standard strategy is thus to optimize $B$ for fixed $A$, then optimize $A$ for fixed $B$, and repeat until convergence. We describe here the proposed optimization algorithm w.r.t. $B$ for fixed $A$ based on \cite{art:ChambollePock11FastPD,art:CombettesPesquet11ProxReview,art:YanezBach14NMF}. The same algorithm can be applied to $A$ for fixed $B$. Let us rewrite problem \eqref{our_model} as:
\vspace{-.1cm}
\begin{eqnarray}
\label{modelFG}
\min_{B\geq0} ~~F(AB) + G(K_B B),
\end{eqnarray}
\vspace{-.1cm}
where
\begin{eqnarray}
&&\!\!\!\!\!\! F(AB) = \textrm{KL}  \big( \Omega \circ ( C\|AB) \big) = \\
&&\!\!\!\!\!\! \sum_{i=1}^m \sum_{j=1}^n  \Big( - \Omega_{ij} C_{ij} \Big( \log \frac{(AB)_{ij}}{C_{ij}} + 1 \Big)  +   \Omega_{ij} (AB)_{ij} \Big), \nonumber\\
&&\!\!\!\!\!\! G(K_B B) = \theta_B \|B\|_{TV_B} = \theta_B  \| K_B B \|_1,
\end{eqnarray}
where $K_B\in\mathbb{R}^{n_e \times m}$ is the graph gradient operator \cite{art:BressonLaurentUminskyVonBrecht13MTV}, with $n_e$ being the number of edges in the graph of $B$. Using the conjugate functions $F^\star$ and $G^\star$ of $F$ and $G$, \eqref{modelFG} is equivalent to the saddle-point problem:
\begin{eqnarray}
\label{modelY1Y2}
\min_{B\geq 0}~~ \max_{Y_1,Y_2}\ ~~ \textrm{tr}( (AB)^T \cdot Y_1) - F^\star (Y_1) + \nonumber \\
\textrm{tr}( (KB^T)^T \cdot Y_2) - G^\star (Y_2),
\end{eqnarray}
where $Y_1\in\mathbb{R}^{n \times m}$, $Y_2\in\mathbb{R}^{n_e \times r}$. Let us now introduce the proximal terms and the time steps $\sigma_1,\sigma_2,\tau_1$, $\tau_2$:
\begin{eqnarray}
\label{modelProx}
&& \min_{B\geq 0}~~ \max_{Y_1,Y_2}\ ~~ \textrm{tr}( (AB)^T \cdot Y_1) - F^\star (Y_1) + \nonumber \\
&& \textrm{tr}( (KB^T)^T \cdot Y_2) - G^\star (Y_2) + \frac{\tau_1 + \tau_2}{2\tau_1\tau_2} \| B - B^k \|_F^2  \nonumber \\
&& - \frac{1}{2\sigma_1} \| Y_1 - Y_1^k \|_F^2 - \frac{1}{2\sigma_2} \| Y_2 - Y_2^k \|_F^2.
\end{eqnarray}
The iterative scheme is thus for $k\geq 0$:
\begin{eqnarray}
Y_1^{k+1} &=& \textrm{prox}_{\sigma_1 F^\star} (Y_1^k + \sigma_1 A B^k),\label{proxF}\\
Y_2^{k+1} &=& \textrm{prox}_{\sigma_2 G^\star} (Y_2^k + \sigma_2 K_B B^k),\label{proxG}\\
B^{k+1} &=& ( B^k - \tau_1 A^T Y_1^{k+1} - \tau_2 ( K_B^T Y_2^{k+1} )^T  )_+,
\end{eqnarray}
where $\textrm{prox}$ is the proximal operator \cite{art:CombettesPesquet11ProxReview} and $(\cdot)_+=\max(\cdot,0)$. For our problem we have chosen the standard Arrow-Hurwicz time steps $\sigma_1 = \tau_1 = 1/\|A\|$ and $\sigma_2 = \tau_2 = 1/\|K\|$, where $\|\cdot\|$ is here the operator norm.

The proximal solutions \eqref{proxF} and \eqref{proxG} are given by:
\begin{eqnarray}
\label{}
\textrm{prox}_{\sigma_1 F^\star} ( Y ) &=& \frac{1}{2} \Big( Y + \Omega - \sqrt{ (Y-\Omega)^2 + 4 \sigma_1 \Omega \circ C }  \Big) \nonumber \\
\label{eq:prox_TV}\textrm{prox}_{\sigma_2 G^\star} ( Y ) &=&  Y - \textrm{shrink}  (Y, \theta_B / \sigma_2 ),
\end{eqnarray}
where $\textrm{shrink}$ is the soft shrinkage operator 
\cite{art:Donoho95SoftThre}.
Note that the same algorithm could be used for Tikhonov regularization, i.e. replacing $\|K_BB\|_1$ by $G(K_B B) = \frac{\theta_B}{2}\|K_B B\|_2^2$ by just changing the first proximal (\ref{eq:prox_TV}) to $\textrm{prox}_{\sigma_2 G^\star} ( Y ) =  \frac{\theta_B}{\sigma_2+\theta_B}Y$.
In \cite{cai2011graph} this regularization is used along with a symmetric version of the KL divergence, however the latter has no analytic solution unlike the one we use in this work. As a result their objective function does not fit an efficient primal dual optimization scheme like the one we propose. We thus choose to keep the non symmetric KL model, denoted as GNMF in this paper, in order to compare the TV versus Tikhonov regularization.


\noindent\\[-7pt]
{\bf Recommending songs.} Once we have learned matrices $A$ and $B$ by solving \eqref{our_model}, we wish to recommend a new playlist $c_{rec}$ given a few songs $c_{in}$ (see Fig. \ref{fig_recom}). We also want to make real-time recommendations, so we design here a fast recommender function as follows: 

Given the songs {$c_{in}$}, we first find a good representation of the query on the learned low-rank space of playlists by solving a regularized least squares problem:\\
$a_{in} = \arg\min_{a\in\mathbb{R}^{1 \times r}} \| \Omega_{in}  \circ (c_{in} - aB )  \|_2^2 + \varepsilon \| a \|_2^2$.
The latter enjoys an analytic solution $a_{in}=( B^T \Omega_{in} B +  \varepsilon I )^{-1}$ $(B^T \Omega_{in} c_{in} )$ that is cheap to compute as $r$ is small (we use $\varepsilon=0.01$). 

The recommended playlist can benefit from the playlists that have similar representations as the one of the query, thus we use the weighted sum $a_{rec}=\sum_{i=1}^n  w_i A_i / \sum_{i=1}^n  w_i$ as the representation of the recommended playlist in the low dimensional space. Here the weights $w_i$ are defined as $w_i=e^{- \| {a_{in}}-A_i\|^2_2/ \sigma^2}$ and depend on the distance of $a_{in}$ from other playlists representations, while $\sigma=\textrm{mean}_i( \{ \| {a_{in}}-A_i \|_2 \}_{i=1}^n)/4$.
The final recommended playlist uses the low-rank representation $a_{rec}$:
\vspace{-.1cm}
\begin{eqnarray}
\label{eq:recommender}
c_{rec} = a_{rec}B.
\end{eqnarray}
\\[-.6cm]
Note finally that the recommended playlist $c_{rec}$ is not binary, but with continued values that serve as song rankings.

\vspace{-6pt}
\section{Graphs of Playlists and Songs}\label{sec:graphs}
\vspace{-6pt}

\textbf{Playlists Graph.}\label{sec:playlist_graph}
The playlists graph naturally encodes pairwise similarities between playlists. The set of nodes of this graph is the set of playlists and the edge weight provides the proximity between two playlists. A large weight (here $w_{ii'}^A\approx 1$) implies a strong proximity between the playlists. In this work, the edge weight of the playlists graph uses both ``outside'' information, i.e. the meta-data, and ``inside'' information, i.e. the songs that form the playlists. As meta-data, we use the predefined Art of the Mix playlist categories \cite{mcfee2012hypergraph} onto which users label their mixes.
The edge weight of the playlists graph is thus defined as follows:

{\vspace{4pt}
\centering{$\displaystyle w_{ii'}^A = \gamma_1 \delta_{cat\{i\}=cat\{i'\}} + \gamma_2 \simty_{\cos}(C_i,C_{i'}),$}\\[4pt]}

\noindent where $cat$ stands for playlist category, $C_i$ is the $i^{th}$ row of matrix $C$ and $\simty_{\cos}(p,q)=p^\top q/(\|p\|.\|q\|)$ is the cosine similarity distance between the vectors of the songs of the two playlists. In our case, the cosine similarity is the ratio between the songs in common and the square root of the product of the lengths of the two playlists. The two positive parameters $\gamma_1,\gamma_2$  with $\gamma_1+\gamma_2=1$ allow to weight the importance of the playlist labels against their element-wise similarity. To control the edge density in each category and to give more flexibility to our recommendation model, we keep a random subset of $20\%$ of the edges between nodes of the same category. As we find experimentally, $\gamma_2 = 0.3$ constitutes a good compromise, see Sec.~\ref{sec:results}.

The quality of the playlist graph is measured by partitioning the graph using the standard Louvain's method \cite{blondel2008fast}. The number of partitions is automatically given by the modularity dendrogram which is cut where the modularity is maximal. The graph used in Sec.~\ref{sec:results} has a modularity of $0.63$ when using the cosine similarity ($\gamma_2 = 0$) only. If we add the meta-data information by connecting $20\%$ of all playlist pairs within each category with $\gamma_2 = 0.3$, the modularity increases to $0.82$.

\noindent\\[-7pt]\textbf{Songs Graph.}\label{sec:song}
The second graph used in our model is the graph of song similarity. It is created from a mixture of Echonest features extracted from the audio signal which we combine with meta-data information and social features for the track. Table~\ref{tab:features_details} gives a view of the features used to create the song graph.

\begin{table}
\centering
\resizebox{\columnwidth}{!}{%
\begin{tabular}{|l |l|}
\multicolumn{2}{l}{{\textbf{High Level Features}}}\\
\hline
acousticness & Acoustic or electric?\\
valence & Is the song positive or negative?\\
energy & How energetic is the song?\\
liveness & Is it a ``live'' recording?\\
speechiness & How many spoken words?\\
danceability & Is the song danceable?\\
tempo & Normalized BPM.\\
instrumentalness & Is the song instrumental?\\
\hline
\noalign{\smallskip}
\multicolumn{2}{l}{{\textbf{Social Features}}}\\
\hline
artist discovery & How unexpectedly popular is the artist?\\
artist familiarity & How familiar is the artist?\\
artist hotttnesss & Is the artist currently popular?\\
song hotttnesss & Is the song currently popular?\\
song currency & How recently has it become popular?\\
\hline
\noalign{\smallskip}
\multicolumn{2}{l}{{\textbf{Temporal Echonest Features}}}\\
\hline
statistics on echonest segments & Described in \cite{schindler2014capturing} \\
\hline
\noalign{\smallskip}
\multicolumn{2}{l}{{\textbf{Metadata Features}}}\\
\hline
genre & ID3 genre extracted from tags given by LastFM api\\
\hline
\end{tabular}
}
\vspace{-7pt}
\caption{The features used to create graph of songs.}
\label{tab:features_details}
\end{table}

In order to improve the quality of our audio features, we trained a Large Margin Nearest Neighbors model \cite{weinberger2005distance} on the song genres extracted from the LastFm associated terms (tags). To extract real music genres we use the Levenshtein distance between those terms weighted by their popularity (according to LastFm) and the music genres defined in the ID3 tags.


Eventually, the songs graph is created using the $k$ nearest neighbors (here $k$ = 5) where the edge weight between two songs $j, j'$ is given by $w_{jj'}^B = \exp(- \norm{x_j - x_{j'}}_1 / \sigma)$ for $j'$ in the $k^{th}$ nearest neighbors of $j$. The parameter $\sigma$ acts as the scale parameter of the graph and is set to be the average distance of the $k^{th}$ neighbors. The obtained graph has a high modularity ($0.64$) and is quite pure with respect to song genres with around 65\% of accuracy using an unsupervised $k$-NN classifier.

\vspace{-6pt}
\section{Experimental Results}\label{sec:results}
\vspace{-6pt}

In this section we validate our approach by comparing our model against three different recommender systems on a real world dataset. Our test dataset is extracted from the Art-of-the-Mix corpus created by McFee and al. in \cite{mcfee2012hypergraph} onto which we extract the previously described features.


Assessing the quality of any music recommender systems is well-known to be a challenging problem \cite{Bonnin:2014ej}. In this work, we use a typical metric for recommender system with implicit feedback, \emph{Mean Percentage Ranking (MPR)} described in \cite{Hu08collaborativefiltering} and the \emph{playlist category accuracy}, that is the percentage of the recommended songs that have already been used in playlists from the requested category in the past.

\noindent\\[-7pt]
\textbf{Models.} We first compare our model against a graphs-only based approach, labeled as \emph{Cosine only}. For a given input, this model computes the $t$-closest playlists (here $t=50$) using cosine similarity. Songs are recommended by computing a histogram of all the songs contained in these playlists weighted by the cosine similarity weight, as defined by eq. \eqref{eq:recommender}. The second model is  NMF using KL divergence, labeled \emph{NMF \cite{art:YanezBach14NMF}}.
The last model, \emph{GNMF  \cite{cai2011graph}} described in Sec.~\ref{sec:models}, is based on the KL divergence with Tikhonov regularization using the graphs of our model.

\noindent\\[-7pt]
\textbf{Queries.} We test our model with three different types of queries. In all cases, a query $c_{test}$ contains $s=3$ input songs, and the system returns the top $k=30$ output songs as a playlist using eq. \eqref{eq:recommender}. The first type of queries, \emph{Random}, contains completely randomly chosen songs from all categories and is solely used as a comparison baseline. 
The second type of queries, \emph{Test}, picks randomly $3$ songs from a playlist of the test set. Lastly, \emph{Sampled}, contains randomly chosen songs from a given category. It simulates a recommender system based on chosen playlist categories input by a user.

\noindent\\[-7pt]
\textbf{Training.} We train our model using a randomly selected subset of $70\%$ of the playlists. 
As our model is not jointly convex, initialization may change the performance of the system, so we use the nowadays standard technique of NNDSVD \cite{boutsidis2008svd} to get a good approximate solution. 
In all our experiments a value of the rank $r=15$ performs well, which is expected as each row has between $5$ and $20$ non-zero values. 
The best set of parameters  $\theta_A = 18$ and $\theta_B = 1$ is found using a grid search using queries on the validation set. In order to prevent overfitting, we perform \emph{early stopping} as soon as the MPR on the validation set ceases to increase. 

\noindent\\[-7pt]
\textbf{Validation set.} We create the ``playlists'' of the validation set by creating artificial queries from the different playlist categories. That is, for each category we randomly pick $s=3$ songs that have been previously used in user-made playlists labeled by the given category. 



\noindent\\[-7pt]
\textbf{Results.} The performance in terms of playlist category accuracy and MPR of the different models are reported in Table~\ref{tab:comparison_pcat} and Table~\ref{tab:comparison_mpr} respectively. As expected, for random category queries all models fail to return playlists from the categories of the input songs. At the same time, the performance of NMF as collaborative filtering without the graphs information is poor. This can be explained by the sparsity of the dataset, that only contains $5$ to $20$ non-zero elements per row, i.e. only 0.11-0.46\% sparsity. Collaborative filtering models are known to perform better as more observed ratings are available \cite{art:KalofoliasBressonBronsteinVandergheynst14MCGraphs}. The cosine model performs better in terms of category accuracy, as it directly uses the cosine distance between the input query and playlists from pure categories. However, its high MPR value shows that our model, albeit more complex, achieves better song recommendations.

\begin{table}[h]
\centering
\begin{tabular}{|l||c|c|c|c|c|}
\hline
{} &  Cosine  & NMF  &  GNMF  & $\gamma_1=0$ & $\gamma_1=0.3$  \\
{} & only &  \cite{art:YanezBach14NMF} & \cite{cai2011graph} & $\gamma_2=1$ &  $\gamma_2=0.7$  \\
\hline
\hline
Random &  0.135 &  0.150 & 0.167 & 0.210 &  0.183 \\
\hline
Test &  0.530 &  0.236 &  0.332 &  0.544 &  \textbf{0.646} \\
Sampled  &  0.822 &  0.237 & 0.366 & 0.598 &  \textbf{0.846} \\
\hline
\end{tabular}
\vspace{-8pt}
\caption{Category accuracy for all models for different types of 3-song queries (higher is better). Results are averaged over 10 train/validation runs with 300 queries each.}
\label{tab:comparison_pcat}
\end{table}

\begin{table}[h]
\centering
\begin{tabular}{|l||c|c|c|c|c|}
\hline
{} & Cosine &  NMF & GNMF & $\gamma_1=0$ & $\gamma_1=0.3$  \\
{} & only &  \cite{art:YanezBach14NMF} & \cite{cai2011graph} & $\gamma_2=1$ &  $\gamma_2=0.7$  \\
\hline
\hline
Test    &  0.208  & 0.248 & 0.181 & 0.153  &  \textbf{0.146}  \\
Sampled &  0.226  & 0.319 &  0.211 & 0.164  &  \textbf{0.074}  \\

\hline
\end{tabular}
\vspace{-8pt}
\caption{Mean percentage ranking (MPR) for all models for different types of 3-song queries (lower is better). Results are averaged over 10 train/validation runs with 300 queries each.}
\label{tab:comparison_mpr}
\vspace{-15pt}
\end{table}

\begin{figure}[h]
 \centerline{
 \includegraphics[width=\columnwidth]{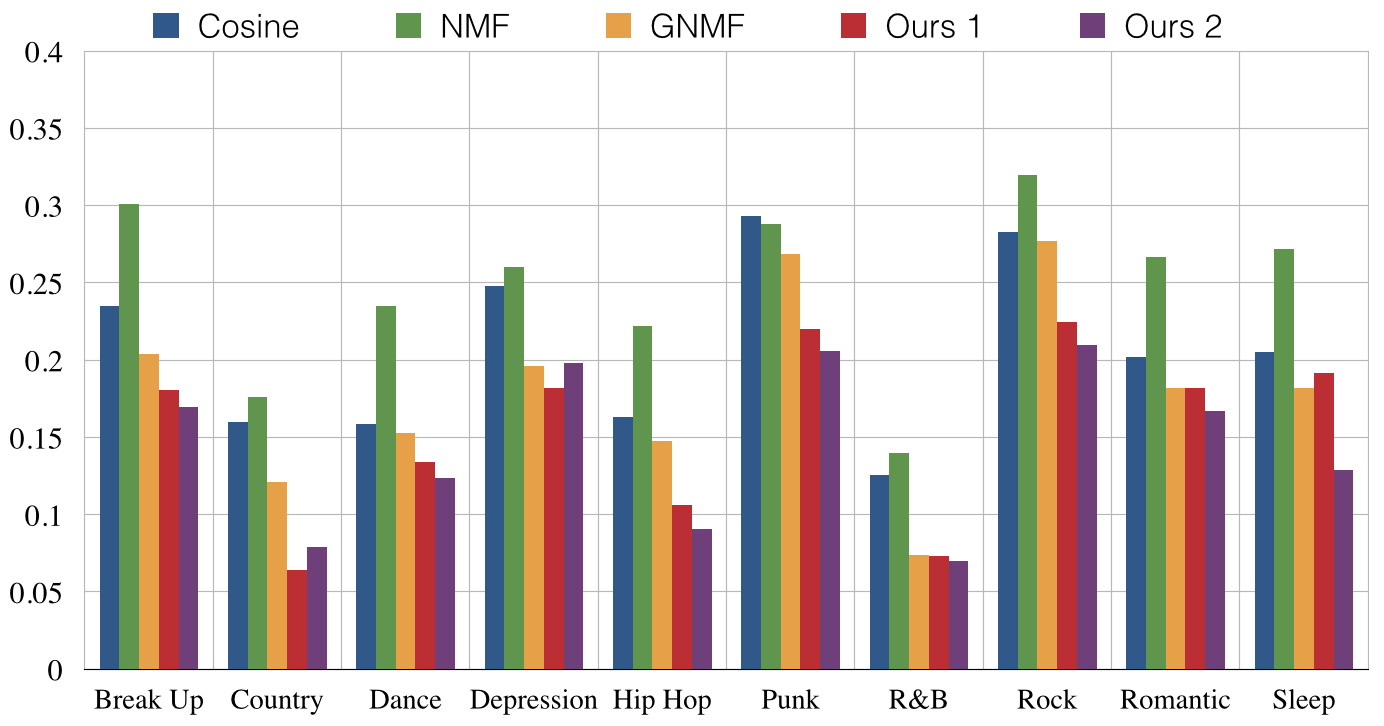}}
 \vspace{-8pt}
 \caption{MPR for each playlist category on the test set. Our models use the same parameters of Table~\ref{tab:comparison_mpr}. Ambiguous categories such as Rock, Punk have the highest MPR on the test set. Our model outperforms significantly the others methods on those specific categories.}
 \label{fig:categories}
\end{figure}




\vspace{-6pt}
\section{Conclusion}\label{sec:conclusion}
\vspace{-6pt}

In this work we introduce a novel flexible song recommender system that combines collaborative filtering with playlist and song proximity information encoded by graphs. We use a primal-dual based optimization scheme to achieve a highly parallelizable algorithm with the potential to scale up to very large datasets. We choose graph TV instead of Tikhonov regularization and demonstrate the model's superiority by comparing our system against three other recommendation models on a real music playlists dataset.


\newpage

\begingroup
\setstretch{0.965}

\begin{thebibliography}{10}

\bibitem{koren2009matrix}
Yehuda Koren, Robert Bell, and Chris Volinsky,
\newblock ``Matrix factorization techniques for recommender systems,''
\newblock {\em Computer}, vol. 8, pp. 30--37, 2009.

\bibitem{Hu08collaborativefiltering}
Yifan Hu, Yehuda Koren, and Chris Volinsky,
\newblock ``Collaborative filtering for implicit feedback datasets,''
\newblock in {\em In IEEE International Conference on Data Mining (ICDM 2008},
  2008, pp. 263--272.

\bibitem{pro:RendleFreudenthalerGantnerSchmidt09Playlist}
S.~Rendle, C.~Freudenthaler, Z.~Gantner, and L.~Schmidt-Thieme,
\newblock ``{BPR: Bayesian Personalized Ranking from Implicit Feedback},''
\newblock in {\em Proceedings of Conference on Uncertainty in Artificial
  Intelligence}, 2009, pp. 452--461.

\bibitem{van2013deep}
Aaron Van~den Oord, Sander Dieleman, and Benjamin Schrauwen,
\newblock ``Deep content-based music recommendation,''
\newblock in {\em Advances in Neural Information Processing Systems}, 2013, pp.
  2643--2651.

\bibitem{shao2009music}
Bo~Shao, Dingding Wang, Tao Li, and Mitsunori Ogihara,
\newblock ``Music recommendation based on acoustic features and user access
  patterns,''
\newblock {\em Audio, Speech, and Language Processing, IEEE Transactions on},
  vol. 17, no. 8, pp. 1602--1611, 2009.

\bibitem{celma2010}
Oscar Celma,
\newblock ``Music recommendation,''
\newblock in {\em Music Recommendation and Discovery}, pp. 43--85. Springer
  Berlin Heidelberg, 2010.

\bibitem{Bonnin:2014ej}
Geoffray Bonnin and Dietmar Jannach,
\newblock ``{Automated Generation of Music Playlists: Survey and
  Experiments},''
\newblock {\em ACM Computing Surveys (CSUR)}, vol. 47, no. 2, pp. 1--35, nov
  2014.

\bibitem{ma2011recommender}
Hao Ma, Dengyong Zhou, Chao Liu, Michael~R Lyu, and Irwin King,
\newblock ``Recommender systems with social regularization,''
\newblock in {\em Proceedings of the fourth ACM international conference on Web
  search and data mining}. ACM, 2011, pp. 287--296.

\bibitem{art:KalofoliasBressonBronsteinVandergheynst14MCGraphs}
V.~Kalofolias, X.~Bresson, M.~Bronstein, and P.~Vandergheynst,
\newblock ``{Matrix Completion on Graphs},''
\newblock {\em arXiv}, vol. preprint arXiv:1408.1717, 2014.

\bibitem{cai2011graph}
Deng Cai, Xiaofei He, Jiawei Han, and Thomas~S Huang,
\newblock ``Graph regularized nonnegative matrix factorization for data
  representation,''
\newblock {\em Pattern Analysis and Machine Intelligence, IEEE Transactions
  on}, vol. 33, no. 8, pp. 1548--1560, 2011.

\bibitem{art:RudinOsherFatemi92ROF}
L.~I. Rudin, S.~Osher, and E.~Fatemi,
\newblock ``{Nonlinear Total Variation Based Noise Removal Algorithms},''
\newblock {\em Physica D}, vol. 60(1-4), pp. 259 -- 268, 1992.

\bibitem{art:CombettesPesquet11ProxReview}
P.L. Combettes and J.C. Pesquet,
\newblock ``{Proximal Splitting Methods in Signal Processing},''
\newblock {\em Fixed-Point Algorithms for Inverse Problems in Science and
  Engineering}, pp. 185--212, 2011.

\bibitem{pro:HaririMobasherBurke12Playlist}
N.~Hariri, B.~Mobasher, and R.~Burke,
\newblock ``{Context-Aware Music Recommendation based on Latent Topic
  Sequential Patterns},''
\newblock in {\em Proceedings of ACM conference on Recommender systems}, 2012,
  pp. 131--138.

\bibitem{lee1999learning}
Daniel~D Lee and H~Sebastian Seung,
\newblock ``Learning the parts of objects by non-negative matrix
  factorization,''
\newblock {\em Nature}, vol. 401, no. 6755, pp. 788--791, 1999.

\bibitem{vavasis2009complexity}
Stephen~A Vavasis,
\newblock ``On the complexity of nonnegative matrix factorization,''
\newblock {\em SIAM Journal on Optimization}, vol. 20, no. 3, pp. 1364--1377,
  2009.

\bibitem{art:YanezBach14NMF}
F.~Yanez and F.~Bach,
\newblock ``{Primal-Dual Algorithms for Non-negative Matrix Factorization with
  the Kullback-Leibler Divergence},''
\newblock {\em arXiv:1412.1788}.

\bibitem{art:BressonLaurentUminskyVonBrecht13MTV}
X.~Bresson, T.~Laurent, D.~Uminsky, and J.H. von Brecht,
\newblock ``{Multiclass Total Variation Clustering},''
\newblock {\em Annual Conference on Neural Information Processing Systems
  (NIPS)}, pp. 1421--1429, 2013.

\bibitem{art:ChambollePock11FastPD}
A.~Chambolle and T.~Pock,
\newblock ``{A First-Order Primal-Dual Algorithm for Convex Problems with
  Applications to Imaging},''
\newblock {\em Journal of Mathematical Imaging and Vision}, vol. 40(1), pp.
  120--145, 2011.

\bibitem{art:Donoho95SoftThre}
D.~Donoho,
\newblock ``{De-Noising by Soft-Thresholding},''
\newblock {\em IEEE Transactions on Information Theory}, vol. 41(33), pp.
  613--627, 1995.

\bibitem{mcfee2012hypergraph}
Brian McFee and Gert~RG Lanckriet,
\newblock ``{Hypergraph Models of Playlist Dialects.},''
\newblock in {\em ISMIR}. Citeseer, 2012, pp. 343--348.

\bibitem{blondel2008fast}
Vincent~D Blondel, Jean-Loup Guillaume, Renaud Lambiotte, and Etienne Lefebvre,
\newblock ``Fast unfolding of communities in large networks,''
\newblock {\em Journal of Statistical Mechanics: Theory and Experiment}, vol.
  2008, no. 10, pp. P10008, 2008.

\bibitem{schindler2014capturing}
Alexander Schindler and Andreas Rauber,
\newblock ``Capturing the temporal domain in echonest features for improved
  classification effectiveness,''
\newblock in {\em Adaptive Multimedia Retrieval: Semantics, Context, and
  Adaptation}, pp. 214--227. Springer, 2014.

\bibitem{weinberger2005distance}
Kilian~Q Weinberger, John Blitzer, and Lawrence~K Saul,
\newblock ``Distance metric learning for large margin nearest neighbor
  classification,''
\newblock in {\em Advances in neural information processing systems}, 2005, pp.
  1473--1480.

\bibitem{boutsidis2008svd}
Christos Boutsidis and Efstratios Gallopoulos,
\newblock ``{SVD Based Initialization: A Head Start For Nonnegative Matrix
  Factorization},''
\newblock {\em Pattern Recognition}, vol. 41, no. 4, pp. 1350--1362, 2008.

\end{thebibliography}

\endgroup

\end{document}